\documentclass[runningheads,a4paper]{llncs}

\usepackage{times}
\usepackage{amssymb}
\usepackage{graphicx}
\usepackage{url}

\usepackage{booktabs}
\usepackage{tabularx}
\usepackage{tcolorbox}
\usepackage{hyperref}
\usepackage[T1]{fontenc} 
\usepackage{lmodern}
\usepackage[autolanguage]{numprint}

\newcommand{\textbluett}[1]{\textcolor{blue}{\texttt{#1}}}
\newcommand{\ms}[1]{{\scriptsize (#1)}}
\newcolumntype{C}{>{\small\centering\arraybackslash}X}

\let\np=\numprint
\newcommand{\keywords}[1]{\par\addvspace\baselineskip
\noindent\keywordname\enspace\ignorespaces#1}

\urldef{\mailsa}\path|{xstano1, hales}@fi.muni.cz|

\begin{document}

\title{Evaluating Prompt-Based and Fine-Tuned Approaches to Czech Anaphora Resolution}

\titlerunning{Evaluating Approaches to Czech Anaphora Resolution}

\author{Patrik Stano \and Ale\v{s} Hor\'ak\orcidID{0000-0001-6348-109X}}
%
%
\authorrunning{P. Stano and A. Hor\'ak}
%
\institute{NLP Centre, Faculty of Informatics, Masaryk University\\
Botanická 68a, Brno, Czech Republic\\ \url{nlp.fi.muni.cz/en}}
%
\index{Stano, Patrik}
\index{Hor\'ak, Ale\v{s}}

\toctitle{} \tocauthor{}

\maketitle

%
%
%
%
\begin{abstract}
Anaphora resolution plays a critical role in natural language understanding, especially in morphologically rich languages like Czech. This paper presents a comparative evaluation of two modern approaches to anaphora resolution on Czech text: prompt engineering with large language models (LLMs) and fine-tuning compact generative models. Using a dataset derived from the Prague Dependency Treebank, we evaluate several instruction-tuned LLMs, including Mistral Large 2 and Llama 3, using a series of prompt templates. We compare them against fine-tuned variants of the mT5 and Mistral models that we trained specifically for Czech anaphora resolution. Our experiments demonstrate that while prompting yields promising few-shot results (up to 74.5\% accuracy), the fine-tuned models, particularly mT5-large, outperform them significantly, achieving up to 88\% accuracy while requiring fewer computational resources. We analyze performance across different anaphora types, antecedent distances, and source corpora, highlighting key strengths and trade-offs of each approach.

\end{abstract}

\section{Introduction}

The use of pronouns to refer to previously mentioned content, a linguistic phenomenon known as anaphora, adds significant complexity to semantic interpretation in natural language processing (NLP). Resolving anaphoric links between pronouns and their antecedents is a crucial step for achieving coherent understanding in tasks such as machine translation, information retrieval, and text generation. Without resolving these references, systems cannot accurately track entities across discourse, which negatively impacts downstream applications.

This challenge is even more pronounced in morphologically rich languages like Czech, where grammatical properties such as case, gender, and number are encoded in word forms. Resolving anaphoric references therefore requires not only syntactic and semantic understanding, but also the ability to handle complex inflectional paradigms. For example, correctly translating a pronoun requires knowing the gender of its antecedent, which may not be straightforward from the surrounding context.

Recent advancements in large language models (LLMs) have shown promise in addressing a wide range of NLP tasks, including those for which they were not explicitly trained. Using prompting strategies, these models can be guided to perform tasks such as question answering and information extraction, as shown in~\cite{agrawal-etal-2022-large}. However, relying on zero-shot or few-shot prompting for anaphora resolution introduces specific limitations. Prompt-based approaches depend heavily on prompt design and model size and are prone to instability or format errors in outputs.

In contrast, fine-tuning generative models, such as mT5~\cite{mt5} or Mistral~\cite{mistral7b}, for specific tasks allows for more controlled behavior and task-specific performance. Although this approach requires annotated data and computational resources for training, it can yield superior results, particularly for structured prediction tasks such as anaphora resolution. In this study, we evaluate the two core approaches, prompt engineering and fine-tuning, on Czech anaphora resolution using a dataset derived from the Prague Dependency Treebank~\cite{pdt}. We aim to compare the strengths, weaknesses, and practical considerations of these two paradigms in the context of a morphologically rich, less-resourced language.

\section{Related Works}

Recent research has explored two main directions for addressing anaphora resolution with neural architectures. The first approach uses generative sequence-to-sequence models such as mT5, a multilingual version of the T5 architecture~\cite{t5}. These models are trained end-to-end to output structured responses based on tagged input sequences. Hicke and Mimno~\cite{hicke2024lions1tigers2} showed that the models can capture coreference relations after fine-tuning the model to generate a copy of an input sentence with annotation tags added to mark the coreference clusters.

The second direction leverages instruction-tuned large language models (LLMs), which can perform complex language understanding tasks through zero- or few-shot prompting. This approach eliminates the need for task-specific training, relying instead on carefully crafted prompts that guide the model to produce the desired output. Studies such as~\cite{le2023largelanguagemodelsrobust} have evaluated this approach for English coreference.

Our work focuses on these two modern paradigms, prompting and fine-tuning, and evaluates their effectiveness for resolving pronominal anaphora in Czech.

\section{Methodology}

\subsection{Dataset}

The dataset used in this study was derived from the Prague Dependency Treebank Consolidated 1.0 (PDTC 1.0)~\cite{pdtc}. It consists of \np{56207} sentences in Czech, each containing one pronominal anaphora and one corresponding antecedent. The sentences are drawn from three major subcorpora: the Prague Discourse Treebank (fiction and newspaper articles), the Prague Translation Corpus (English-Czech translations), and the Prague Spoken Corpus (transcripts of spoken Czech).

Each sentence is annotated with a single anaphor-antecedent pair using XML-style tags: \texttt{\textcolor{blue}{<ana>}\textcolor{blue}{</ana>}} for the anaphor and \texttt{\textcolor{blue}{<ant>}\textcolor{blue}{</ant>}} for the antecedent. In addition, metadata is provided for each example, including the type of coreference (grammatical or textual), the token-level distance between the anaphor and antecedent, the subcorpus of origin, and the morphosyntactic category of the pronoun. \autoref{fig:gram_text_anaphora} presents examples of the two anaphora types in the dataset. In the grammatical anaphora sentence, the pronoun \textit{\textbluett{<ana>}which\textbluett{</ana>}} refers to the subject ``\texttt{budova}'' (\emph{building}), and the link is established primarily through syntactic means, which is characteristic of grammatical coreference, i.e.\ a semantic context is not needed. In the textual anaphora example, discerning what the pronoun \textit{\textbluett{<ana>}him\textbluett{</ana>}} refers to requires semantic understanding of the context, which characterizes textual anaphora.

\begin{figure}[t]
    \textbf{Grammatical anaphora:}

    \begin{tcolorbox}[colback=gray!10, colframe=black, width=\textwidth, sharp corners, boxrule=0.5pt]\footnotesize\raggedright
    \texttt{Budova, \textbluett{<ana>}která\textbluett{</ana>} byla dokončena v loňském roce, stále nebyla otevřena.}

    \emph{The building, \textbluett{<ana>}which\textbluett{</ana>} was completed last year, is still not open.}
    \end{tcolorbox}

    \vspace{0.5em}

    \textbf{Textual anaphora:}

    \begin{tcolorbox}[colback=gray!10, colframe=black, width=\textwidth, sharp corners, boxrule=0.5pt]\footnotesize\raggedright
    \texttt{Tomáš se domluvil s Jardou, že \textbluett{<ana>}ho\textbluett{</ana>} odveze na nádraží.}

    \emph{Tomáš agreed with Jarda to take \textbluett{<ana>}him\textbluett{</ana>} to the station.}
    \end{tcolorbox}

    \caption{Examples of Grammatical and Textual Anaphora in the
    dataset.}
    \label{fig:gram_text_anaphora}
\end{figure}

To ensure clarity and alignment with human annotation quality, several constraints were applied during data extraction: only pronominal anaphora were considered, each text passage was limited to a maximum of three sentences of context, and only pairs with explicit referential links were included. These decisions help standardize the learning task and reduce ambiguity for both prompt-based and fine-tuned models.

\begin{table}[b]
    \caption{Statistics of the evaluation dataset. The numbers $X/Y$
    represent the counts of grammatical/textual anaphora.}
    \label{tab:dataset}
  \begin{tabularx}{\textwidth}{XCCC|C}
    \toprule
    & Train & Validation & Test & Total\\
    \midrule
    text passages & \np{25951}/\np{19009} & \np{3244}/\np{2376} & \np{3247}/\np{2380} & \np{32442}/\np{23765} \\
    sentences & \np{34413}/\np{30757} & \np{4369}/\np{3913} & \np{4351}/\np{3837} & \np{43133}/\np{38507} \\
    words & \np{750298}/\np{601294} & \np{93324}/\np{74365} & \np{95363}/\np{75385} & \np{938985}/\np{751044} \\
    \bottomrule
  \end{tabularx}
\end{table}

The dataset is split into training, validation, and test subsets with the content presented in~\autoref{tab:dataset}. The data is available for download in the Hugging Face repository.\footnote{\url{https://huggingface.co/datasets/MU-NLPC/pdt_anaphora_czech}}

\subsection{Experiments}

In our extensive evaluation of current anaphora resolution approaches,
we have compared prompt engineering using multilingual LLMs and fine-tuning selected encoder-decoder and decoder-only models.

\subsubsection{Prompt Engineering.}

Within this category, six instruction-tuned LLMs were evaluated -- Mistral Large 2 \ms{123B}~\cite{mistral_large}, LLaMA 3.1 \ms{70B}~\cite{llama3}, LLaMA 3.2 \ms{3B}~\cite{llama3}, Aya 23 \ms{35B}~\cite{aya23}, Gemma 2 \ms{27B}~\cite{gemma2}, and Mistral (v0.2, 7B)~\cite{mistral7b}. The task prompts were presented in English as this did not impact the performance of larger models while slightly improving the performance of smaller ones and were tested in zero-shot, one-shot, and three-shot settings. Three main prompting strategies were evaluated:
\begin{itemize}
\item \textbf{Yes/No classification:} The model is asked whether a given antecedent matches the pronominal anaphor, see~\autoref{fig:yesnoprompt}.

\begin{figure}[t]
    \centering
    \begin{tcolorbox}[colback=gray!10, colframe=black, width=\textwidth, sharp corners, boxrule=0.5pt]
        \textbf{Prompt:}
\begin{verbatim}
    You are an anaphora resolution system. In the following
    sentence: "$sentence_ana$" does "$anaphora$" refer to
    "$antecedent_subtree$" ? Respond only YES or NO. Do not
    include anything else in your response.
\end{verbatim}
    \end{tcolorbox}
    \caption{Prompt template for the Yes/No experiment.}
    \label{fig:yesnoprompt}
\end{figure}

\item \textbf{Question-answering:} The model is instructed to return the antecedent span in response to a question. This experiment is directly inspired by~\cite{le2023largelanguagemodelsrobust}, specifically the question-answering template, which involves prompting the model with an excerpt in which the anaphor is highlighted and posing an open-ended question without candidate mentions provided as in~\autoref{fig:questionprompt}.

\begin{figure}[t]
    \centering
    \begin{tcolorbox}[colback=gray!10, colframe=black, width=\textwidth, sharp corners, boxrule=0.5pt]
        \textbf{Prompt:}\\
        \textbf{INSTRUCTIONS:} You are an anaphora resolution system. You are given a sentence in which a word is marked with \textbluett{<ana></ana>} tags. Your task is to identify which passage of the sentence the mention marked in \textbluett{<ana></ana>} refers to. Answer in format [X] where X is the passage of the sentence that the marked mention refers to. Do not change the grammatical form of this passage. Do not include anything else in your answer.
                \begin{tcolorbox}[colback=gray!5, colframe=gray!10, sharp corners, boxrule=0pt, left=0mm, right=0mm, top=0mm, bottom=0mm]
            \textbf{SENTENCE:} "$sentence\_ana$"\\
\textbf{QUESTION:} Which passage of the sentence does \textbluett{<ana>}$anaphora$\textbluett{</ana>} refer to? Answer in the format as instructed\\
\textbf{ANSWER:} [$antecedent\_subtree$]
        \end{tcolorbox}
\textbf{SENTENCE:} "$sentence\_ana$"\\
\textbf{QUESTION:} Which passage of the sentence does \textbluett{<ana>}$anaphora$\textbluett{</ana>} refer to? Answer in the format as instructed\\
\textbf{ANSWER:} 
    \end{tcolorbox}
    \caption{Prompt template for the question-answering experiment. The highlighted section shows how the few-shot prompting was implemented.}
    \label{fig:questionprompt}
\end{figure}

\item \textbf{Tagging:} The model must insert \textbluett{<ant>}\textbluett{</ant>} tags around the corresponding antecedent in the sentence as instructed in~\autoref{fig:tagprompt}.

\begin{figure}[t]
    \centering
    \begin{tcolorbox}[colback=gray!10, colframe=black, width=\textwidth, sharp corners, boxrule=0.5pt]
        \textbf{INSTRUCTIONS:} You are an anaphora resolution system. You are given a sentence in which a word is marked with \textbluett{<ana></ana>} tags. Your task is to identify which passage of the sentence the mention marked in 
        \textbluett{<ana></ana>} refers to. Add \textbluett{<ant></ant>} tags to the sentence around the part of the sentence that the mention marked in 
        \textbluett{<ana></ana>} refers to. Answer in format [X] where X is the original sentence with the \textbluett{<ant></ant>} tags added. Do not include anything else in your answer.
                \begin{tcolorbox}[colback=gray!5, colframe=gray!10, sharp corners, boxrule=0pt, left=0mm, right=0mm, top=0mm, bottom=0mm]
            \textbf{SENTENCE:} "$sentence\_ana$"\\
\textbf{QUESTION:} Which passage of the sentence does \textbluett{<ana>}$anaphora$\textbluett{</ana>} refer to? Answer in the format as instructed\\
\textbf{ANSWER:} [$sentence\_ant\_ana$]
        \end{tcolorbox}
\textbf{SENTENCE:} "$sentence\_ana$"\\
\textbf{QUESTION:} Which passage of the sentence does \textbluett{<ana>}$anaphora$\textbluett{</ana>} refer to? Answer in the format as instructed\\
\textbf{ANSWER:} 
    \end{tcolorbox}
    \caption{Prompt template for the tagging experiment. The highlighted section shows how the few-shot prompting was implemented.}
    \label{fig:tagprompt}
\end{figure}

\end{itemize}

\subsubsection{Fine-Tuning.}

We fine-tuned four models on the training set -- mT5-small, mT5-base, mT5-large, and Mistral 0.2. For mT5, we used a sequence-to-sequence format where the input includes the sentence with anaphora marked and the target adds the \textbluett{<ant>}\textbluett{</ant>} tags. For Mistral, we used LoRA (Low-Rank Adaptation~\cite{hu2022lora}) for parameter-efficient fine-tuning. The training hyperparameters included the learning rate of 2e--5, batch size of 1 for mT5, and a length constraint for decoding outputs.

\subsection{Evaluation Metrics}
\label{sec:evaluation-metric}

The model predictions were evaluated using a relaxed span-level accuracy measure tailored for the structure of the task. Each input sentence contains one pronominal anaphor marked by \textbluett{<ana>}\textbluett{</ana>} tags.
A prediction is considered correct if the model correctly identifies a corresponding \textbluett{<ant>}\textbluett{</ant>} span that meets the following criteria:
\begin{enumerate}
\item \textbf{Root token match:} The syntactic root of the gold antecedent span (marked \texttt{an\-te\-ced\-ent\_root} in the dataset) must be included in the predicted span.
\item \textbf{Span containment:} The predicted span must be fully contained within the token boundaries of the gold antecedent span (marked \texttt{antecedent\_subtree} in the dataset).
\item \textbf{One-to-one structure:} The predicted antecedent span must be a continuous section of the input.
\end{enumerate}
This relaxed evaluation avoids penalizing variations in specificity, while ensuring that the essential coreference relation is captured. For example both ``\texttt{strom}'' (\emph{tree}) and ``\texttt{šťastný strom rostoucí v lese}'' (\emph{happy tree growing in a forest}) may be valid antecedents for ``\texttt{na něj}'' (\emph{at it}) in the context. Penalizing the former for being too short or the latter for being too descriptive would obscure the model's correct understanding of coreference. The metric is suitable for both prompt-based and generative models, which may produce variable surface forms even when the reference is correctly understood.

\subsection{Baseline}

To contextualize the performance of prompt-based and fine-tuned models, we implemented a rule-based baseline. The baseline identifies the antecedent based on syntactic and surface-level heuristics, without access to gold annotations or training data.
Given a sentence containing a single anaphor tag \textbluett{<ana>}\textbluett{</ana>}, the system first extracts all noun phrases (NPs) that appear before the anaphor. The NP boundaries and syntactic heads are identified by UDPipe~\cite{straka2018udpipe}. Among these candidates, the closest preceding NP that matches the pronoun in gender and number (as determined by the morphological features predicted by UDPipe) is selected. 
The baseline score reported uses the relaxed accuracy metric (as defined in \autoref{sec:evaluation-metric}): the predicted span includes the syntactic root of the gold antecedent and lies within its token span.

This rule-based approach serves as a lower bound for the task performance and provides insight into how much improvement is achieved through learned models.

\section{Results}

The following section presents the results achieved by both approaches. The prompting experiments were conducted using an NVIDIA A40 graphics card for each model, except for Mistral large 2 \ms{123B} and Llama 3.1 \ms{70B}, which used an NVIDIA A100. An NVIDIA A40 was also used for fine-tuning.

\subsection{Prompting Results}

\begin{table}[bt]
  \caption{Question-answering experiment results (accuracy)}
  \label{tab:questiondirect}
  \begin{tabularx}{\textwidth}{XCCC}
    \toprule
    Model & Zero-shot & One-shot & Three-shot \\
    \midrule
    Rule-Based baseline & 0.310 \\
    Mistral large 2 \ms{123B}\hspace*{-7pt} & $0.695$ & $0.734$ & $0.745$\\
    Llama 3.1 \ms{70B} & $0.486$ & $0.553$ & $0.561$\\
    Aya 23 \ms{35B} & $0.139$ & $0.534$ & $0.563$ \\
    Gemma 2 \ms{27B} & $0.436$ & $0.537$ & $0.573$ \\
    Mistral 0.2 \ms{7B} & $0.247$ & $0.314$ & $0.296$ \\
    Llama 3.2 \ms{3B} & $0.014$ & $0.063$ & $0.045$ \\
    \bottomrule
  \end{tabularx}
\end{table}

The prompting experiments revealed that the resulting performance is heavily influenced by both the model size and the prompt formulation. Among the tested models, Mistral Large 2 achieved the highest accuracy in the question-answering setup with 74.5\% accuracy in the three-shot setting. Other large models such as LLaMA 3.1 70B and Aya 23 also showed moderate performance, but smaller models struggled significantly.
\autoref{tab:questiondirect} summarizes the best accuracy scores for each model in the question-answering setup across zero-, one-, and three-shot scenarios. The results mostly fit the conversational nature of the models, meaning they are able to display the best performance when a task is framed using an open-ended question.

\autoref{tab:yesno} offers the results of the Yes/No experiment. With less specificity, the scores fit the hypothesis of representing the upper limits for the model performance. In each case, the model score in the question-answering experiment was greater than in the other two experiments, in some cases by a large margin.

As shown in \autoref{tab:tag}, each model experienced a significant drop in performance in the third experiment. The purpose here was to achieve the same output formatting as is used in the fine-tuning (without seeing the training data). Since the task is not notably different from the question-answering experiment, accuracies could be improved by structured prompting, which involves splitting the task into multiple steps. However, the increased length of the generated outputs caused by generating the entire sentence rather than just the antecedent significantly increased processing time. This is the reason why the largest model, Mistral large 2, was not evaluated in this experiment.

\begin{table}[t]
  \caption{Yes/No experiment results (accuracy)}
  \label{tab:yesno}
  \centering
  \begin{tabularx}{.7\textwidth}{XC}
    \toprule
    Model & Zero-shot  \\
    \midrule
    Mistral large 2 \ms{123B} & 0.865 \\
    Llama 3.1 \ms{70B} & 0.809 \\
    Aya 23 \ms{35B} & 0.992 \\
    Gemma 2 \ms{27B} & 0.965 \\
    Mistral 0.2 \ms{7B} & 0.907 \\
    Llama 3.2 \ms{3B} & 0.602 \\
    \bottomrule
  \end{tabularx}
\end{table}

\begin{table}[bt]
  \caption{Tag experiment results (accuracy)}
  \label{tab:tag}
  \begin{tabularx}{\textwidth}{XCCC}
    \toprule
    Model & Zero-shot & One-shot & Three-shot \\
    \midrule
    Rule-Based baseline &
    0.310 \\
    Llama 3.1 \ms{70B} & $0.329$ & $0.335$ & $0.290$\\
    Aya 23 \ms{35B} & $0.015$ & $0.043$ & $0.074$ \\
    Gemma 2 \ms{27B} & $0.131$ & $0.072$ & $0.1$ \\
    Mistral 0.2 \ms{7B} & $0.033$ & $0.111$ & $0.121$ \\
    Llama 3.2 \ms{3B} & $0.018$ & $0.024$ & $0.023$ \\
    \bottomrule
  \end{tabularx}
\end{table}

\subsection{Fine-Tuning Results}

The fine-tuned models demonstrated much stronger performance than the prompted ones. The accuracies achieved by each model are shown in \autoref{tab:finetuned}. The best result was obtained using mT5-large, which achieved 88\% accuracy on the test set. Even the compact mT5-base model outperformed all prompted models, suggesting that the task-specific fine-tuning is more effective than zero- or few-shot prompting. The result of the Mistral 0.2 model could possibly be improved by further training for more epochs, although in our measurements increasing the number of epochs from 3 to 6 achieved only a moderate accuracy improvement of 0.03 points. 

Each model performed better on grammatical anaphora than on textual anaphora -- for example, the average mT5-large accuracy of 0.880 splits to 0.917 for grammatical anaphoras and 0.829 for textual anaphoras. This result is likely based on the fact that grammatical anaphoras are generally simpler and follow regular patterns, which are well represented in the training dataset. Textual anaphora require context, and additionally do not follow syntactic patterns, making it more difficult to cover their variations in the dataset.
As \autoref{tab:mt5-large} shows, the accuracy of the mT5-large model depended heavily on the pronoun type of the anaphora being resolved. While the model achieved high accuracy on both types of grammatical anaphora (indefinite pronouns and definite personal pronouns), the model only achieved an accuracy higher than 0.9 on textual anaphora when the referring pronoun was a definite personal pronoun.

\begin{table}[t]
  \caption{Fine-tuned models performance}
  \label{tab:finetuned}
  \centering
  \begin{tabularx}{.7\textwidth}{XC}
    \toprule
    Model & Accuracy  \\
    \midrule
    Rule-Based baseline &
    0.310 \\
    mT5-small \ms{300M} & 0.660 \\
    mT5-base \ms{580M} & 0.831 \\
    mT5-large \ms{1.2B} & 0.880 \\
    Mistral 0.2 \ms{7B} & 0.787 \\
    \bottomrule
  \end{tabularx}
\end{table}

\begin{table}[bt]
  \caption{mT5-large accuracy by pronoun category and anaphora type (grammatical anaphora with a referring demonstrative pronoun do not occur)}
  \label{tab:mt5-large}
  \begin{tabularx}{\textwidth}{p{3.5cm}XCC}
    \toprule
     & Grammatical & Textual & Average \\
    \midrule
    n.pron.indef & $0.909$ & $0.714$ & $0.907$\\
    n.pron.def.pers & $0.939$ & $0.901$ & $0.915$ \\
    n.pron.def.demon & $-$ & $0.677$ & $0.677$ \\
    \bottomrule
  \end{tabularx}
\end{table}

We believe that the lower accuracy for textual anaphora with referring indefinite pronouns is misleading and should not be considered an accurate representation of the model's abilities since it is a very specific category. The dataset contains only 192 examples of this type and only 21 of these are part of the test subset. This category mostly contains uncommon, outlier sentences with a complicated structures where the antecedent would be difficult for even a human evaluator to identify. 

The lowest accuracy of 0.677 was achieved with textual anaphora and referring demonstrative pronouns. In this case, the reason does not lie in their irregularity or insufficient representation in the training split. This type of anaphora is characterized by the antecedent being a whole phrase rather than an entity. This makes the anaphora more complex and difficult to evaluate.

\begin{table}[bt]
  \caption{mT5-large accuracy by antecedent distance}
  \label{tab:mt5-large_distance}
  \centering
  \begin{tabularx}{.7\textwidth}{XC}
    \toprule
    Distance & Accuracy \\
    \midrule 
    Cataphora & $0.828$\\
    Anaphor in antecedent & $0.713$\\
    0-5 & $0.899$ \\
    6-10 & $0.871$ \\
    11-20 & $0.852$ \\
    21-30 & $0.874$ \\
    \bottomrule
  \end{tabularx}
\end{table}

Examining the relationship between the accuracy and the distance between antecedents and anaphors, as shown in \autoref{tab:mt5-large_distance}, reveals no trend suggesting that the accuracy decreases with an increasing distance. Anaphora is a localized phenomenon in order to be easily understandable. The context length of current LLMs exceeds the length of the span between an antecedent and an anaphor by orders of magnitude, which explains this result. Accuracy significantly decreased in cases where the anaphor was part of the antecedent. This is likely due to the higher complexity of these sentences. The mT5 models are bidirectional transformers, so the fact that a part of the antecedent comes after the anaphor should not negatively affect accuracy. This can also be seen by comparing the accuracy achieved with cataphoric, i.e.\ forward directing, references, where a decrease was observed, but not to the extent of the anaphor-in-antecedent cases.

Finally, as shown in~\autoref{tab:mt5-large_corpus}, the model achieved similar accuracy on examples of original Czech text and text translated from English. The lower accuracy achieved on spoken data may be due to its underrepresentation in the training set (13.9\%) as well as the fact that spoken communication is often less structured and formal.

\begin{table}[bt]
  \caption{mT5-large accuracy by subcorpus}
  \label{tab:mt5-large_corpus}
  \centering
  \begin{tabularx}{.9\textwidth}{XC}
    \toprule
    Subcorpus & Accuracy \\
    \midrule 
    PDT 3.5 (original Czech) & $0.889$\\
    PCEDT 2.0 (English translated) & $0.885$ \\
    PDTSC 2.0 (Spoken Czech) & $0.835$ \\
    \bottomrule
  \end{tabularx}
\end{table}

When comparing the inner results achieved by both main approaches, prompting and fine-tuning, there are no significant differences in their strengths and weaknesses across categories. Textual anaphora proved more difficult to resolve for models fine-tuned for the task as well as for models that were not. This can be seen by comparing the fine-tuned results to the results of Mistral large 2, see \autoref{tab:mt5-mistral-anatype}, achieved by prompting. The same is true for the anaphora pronoun types. The relative results of the fine-tuning approach, presented in \autoref{tab:mt5-large}, are mirrored by the results of prompting. 
This comparison has an important implication. It shows that the variance in the accuracy of the fine-tuned models is likely not based on imbalances in the training data. Even models that were not specifically fine-tuned for the task, i.e.\ not exposed to the dataset, demonstrate the same variance.  

\begin{table}[bt]
  \caption{Fine-tuned mT5-large and prompted Mistral large 2 accuracies by anaphora type}
  \label{tab:mt5-mistral-anatype}
  \begin{tabularx}{\textwidth}{XCCC}
    \toprule
    & Grammatical & Textual & Average \\
    \midrule 
    mT5-large \ms{1.2B} & $0.917$ & $0.829$ & $0.880$ \\
    Mistral large 2 \ms{123B} & $0.782$ & $0.695$ & $0.745$ \\
    \bottomrule
  \end{tabularx}
\end{table}

\section{Discussion}

The results of our experiments indicate that fine-tuning a task-specific model is more effective for Czech anaphora resolution than relying on prompting with large multilingual LLMs. While LLMs such as Mistral Large 2 can approach usable performance levels with carefully designed prompts, even moderately sized fine-tuned models like mT5-base consistently outperform them.

These findings highlight a key trade-off between flexibility and performance. Prompting offers a low-barrier solution that does not require training infrastructure or labeled data, making it attractive for rapid prototyping. However, this convenience comes at the cost of lower performance, limited control over outputs, and high resource requirements for inference with large models.

Fine-tuning, on the other hand, delivers better accuracy and generalization, even with smaller models. The mT5-large model not only achieved the best results overall, but also demonstrated robustness across different types of anaphora and domains (e.g., fiction, news, spoken language). Fine-tuned models are also easier to integrate into pipelines requiring structured output.

The observed performance gap between prompt-based and fine-tuned systems is especially relevant in low-resource language settings like Czech, where computational efficiency and domain adaptability are critical. Our findings suggest that investing in annotated data and fine-tuning smaller models remains a competitive strategy despite recent popularity of increasingly large models and prompting approaches.

Nevertheless, prompt-based approaches remain valuable in scenarios where labeled data is scarce or inference flexibility is paramount. Improvements in prompt design, instruction tuning, and output postprocessing may help reduce the current performance gap in future work.

\section{Conclusions and Future Directions}

This work has presented an exploration of anaphora resolution techniques with a focus on the Czech language, examining both fine-tuning and prompt engineering approaches.
While the prompt engineering approach proved feasible, particularly in a few-shot context, it was less efficient and accurate compared to the fine-tuned models, especially as the task complexity increased. The models achieved the highest accuracy when prompted in a conversational format. The findings suggest that, even for languages with limited annotated data, such as Czech, fine-tuning remains a more practical and effective method for anaphora resolution. On the other hand, the experiments conducted suggest that the current large language models have reasonable anaphora resolution capabilities even without being explicitly trained for the task. 
In the future work, we plan to extend this research by including a mention detector in the pipeline to convert the fine-tuned model into a coreference resolution solution. Other directions worth exploring are the inclusion of multilingual data into the training process and employing a mixture of experts approach to divide the anaphora resolution model into expert networks along with a pronoun category classifier.

\subsection*{Acknowledgments}
This work has been partly supported by the Ministry of Education, Youth and Sports of the Czech Republic within the LINDAT-CLARIAH-CZ project LM2023062.
The authors acknowledge the OSCARS project, which has received funding from the European Commission's Horizon Europe Research and Innovation programme under grant agreement No. 101129751.
Computational resources were provided by the e-INFRA CZ project (ID:90254), supported by the Ministry of Education, Youth and Sports of the Czech Republic.

\bibliographystyle{splncs04}
\bibliography{samplepaper}

\end{document}